\documentclass[10pt,twocolumn,twoside]{IEEEtran}

\usepackage{amsmath}
\usepackage{amssymb}
\usepackage{graphics}
\usepackage{epstopdf}
\usepackage{amsmath}
\usepackage{algorithmicx}
\usepackage{algpseudocode}
\usepackage{algorithm}
\usepackage{multirow}
\usepackage{bm}
\usepackage{enumerate}
\usepackage{caption, subcaption}
\usepackage{paralist}
\usepackage{color}
\usepackage{dsfont}
\usepackage{array}

\usepackage{xfrac}
\usepackage{multirow}
\usepackage{verbatim}

\usepackage[english]{babel}

\usepackage{color}
\definecolor{purple(x11)}{rgb}{0.63, 0.36, 0.94}
\definecolor{cadmiumgreen}{rgb}{0.0, 0.42, 0.24}

\begin{document}

\title{Fast Orthonormal Sparsifying Transforms Based on Householder Reflectors}
\author{Cristian Rusu\thanks{C. Rusu is with the Institute for Digital Communications, School of Engineering, The University of Edinburgh. N. Gonz\'{a}lez-Prelcic is with the Atlantic Research Center for Information and Communication Technologies, University of Vigo. R. Heath is with The University of Texas at Austin.


C. Rusu (c.rusu@ed.ac.uk) is the corresponding author. Demo source code available online at https://udrc.eng.ed.ac.uk/sites/udrc.eng.ed.ac.uk/files/attachments/demo.zip.

C. Rusu acknowledges support from the Engineering and Physical Sciences Research Council (EPSRC) [EP/K014277/1] and the MOD University Defence Research Centre (UDRC) on Signal Processing. N. Gonz\'{a}lez-Prelcic acknowledges support from the Spanish
Government and the European Regional Development Fund (ERDF) under project TACTICA. R. Heath acknowledges support from the National Science Foundation under Grant No. NSF-CCF-1319556.

}, Nuria Gonz\'{a}lez-Prelcic and Robert W. Heath Jr.}

\maketitle

\begin{abstract}
Dictionary learning is the task of determining a data-dependent transform that yields a sparse representation of some observed data. The dictionary learning problem is non-convex, and usually solved via computationally complex iterative algorithms. Furthermore, the resulting transforms obtained generally lack structure that permits their fast application to data. To address this issue, this paper develops a framework for learning orthonormal dictionaries which are built from products of a few Householder reflectors. Two algorithms are proposed to learn the reflector coefficients: one that considers a sequential update of the reflectors and one with a simultaneous update of all reflectors that imposes an additional internal orthogonal constraint. The proposed methods have low computational complexity and are shown to converge to local minimum points which can be described in terms of the spectral properties of the matrices involved. The resulting dictionaries balance between the computational complexity and the quality of the sparse representations by controlling the number of Householder reflectors in their product. Simulations of the proposed algorithms are shown in the image processing setting where well-known fast transforms are available for comparisons. The proposed algorithms have favorable reconstruction error and the advantage of a fast implementation relative to the classical, unstructured, dictionaries.
\end{abstract}


\section{Introduction}


Sparsifying transforms \cite{SparseRep} allow efficient representation of data when a data-dependent overcomplete dictionary is available. Overcomplete dictionaries are useful in image processing \cite{DictImageDenosing, DictAstronomicalImageDenoising, DictDeblurr}, speech processing \cite{Speech1} and wireless communications \cite{Ayach_TWC14, OurICC}. Unfortunately, the selection of a sparsifying transform involves solving a non-convex optimization problem for a dictionary matrix $\mathbf{D}$ such that a real data set can be represented with a sparse representation matrix $\mathbf{X}$ whose sparsity level is constrained. Because direct solution of the optimization method is difficult \cite{SparseIsNPHard, DictLearningIsNPHard}, proposed algorithms seek a suboptimal solution via alternating minimization.

Most prior work considers alternating minimization for dictionaries that are overcomplete. Algorithms like the method of optimal directions (MOD) \cite{MOD}, K-SVD \cite{KSVD} and algorithms based on direct optimization \cite{DirectDictionary} all perform alternating updates, differing in the ways they actually perform the update of the dictionary and of the sparse representations. Unfortunately, most general solutions to the dictionary learning problem are relatively slow in computing the dictionary and they lack a formal analysis of performance. Some of these difficulties stem from the fact that the proposed algorithms produce non-orthonormal, even overcomplete, dictionaries. Furthermore, overcomplete transforms themselves present some disadvantages when compared to the classical, fixed and fast, transforms. In any application, one general drawback of these computed dictionaries is that they need to be stored (or transmitted) along with the encoded/compressed data. Another, and more important, drawback is that representing vectors in a non-orthonormal or overcomplete dictionary involves a non-linear, computationally expensive, procedure \cite{GreedIsGood, JustRelax}.


Fast transforms allow more efficient application of the dictionary to compute the sparse representation. For example, the discrete cosine, Fourier, Hadamard or wavelet transforms all have computationally efficient implementations, i.e., for example $O(n\log n)$ computational complexity \cite{FFT}. These fast transforms are widely used in signal and image processing but unfortunately are not the best sparsifying transforms in every situation. 

Recent work has devised fast sparsifying dictionaries that are built from fast transforms. For example, one of the first proposed algorithm called sparse K-SVD \cite{DoubleSparsity}, considers constructing a dictionary by using sparse linear combinations of the components of a fast transform. These dictionaries are efficient to apply since a linear combination of just a few components (which themselves are computed fast) can be done efficiently. The second, more recent, approach \cite{EfficientDictionaries} considers factorizing the dictionary as a product of a few very sparse matrices that can be easily manipulated. This is in the spirit of several fixed sparsifying transform that have this property, like the aforementioned Hadamard case which enjoys a factorization as a product of sparse matrices. Other approaches, like the one in \cite{DictCirculant} treats each atom of the dictionary as the composition of several circular convolutions so that the overall dictionary can be manipulated quickly, via Fourier transforms. The approach in \cite{Rusu2013} is to construct an overcomplete dictionary from concatenations of several orthonormal sub-dictionaries and partition the sparse representations such that they belong exclusively to only one sub-dictionary. Tree structures have been used to quickly constructing sparse approximations \cite{QuickSparse}. The learning algorithms proposed in \cite{DoubleSparsity}--\cite{QuickSparse} are slow in general, lack performance analysis or guarantees and usually involve relatively complex algorithms and extra data structures for the description of the dictionary. 
The approach in \cite{ShiftInvariantDictLearning}, provides a fast procedure for learning circulant dictionaries but, unfortunately, these dictionaries are not a general solution due to their low number of degrees of freedom.

In this paper we develop algorithms for finding orthonormal dictionaries that can be used directly and inversely faster than the unconstrained, general, orthonormal dictionaries. We reduce the computational complexity of manipulating the dictionaries by considering that they are products of only a few Householder reflectors \cite{Golub1996}. While any orthonormal dictionary of size $n \times n$ can be factorized into $n$ reflectors, in this paper we use $m \ll n$ reflectors in the structure of the dictionary. This way, by applying the reflectors sequentially, low complexity dictionary manipulation is achieved. We choose to use Householder reflectors as the building blocks of our dictionaries since they enjoy low complexity manipulation, e.g., the reflector-vector product is computed in $O(n)$. By using fewer reflectors than needed, our algorithms cannot explore the entire space of orthonormal dictionaries rather only a subset of these. The main advantage though is the low complexity manipulation of the dictionaries designed this way. In general, an open question is if all orthonormal and Hessian matrices can be well represented and approximated with low complexity \cite{FastApproximations} (factored into $(1/2) n \log n$ Givens rotations).

In this paper, we propose two algorithms that compute the coefficients of the Householder reflectors. The first approach builds an orthonormal dictionary composed of just a few reflectors by updating all the coefficients of each reflector sequentially, keeping the other ones fixed. The main advantages of this approach are: (i) each reflector update is done efficiently by solving an eigenvalue problem and (ii) the overall performance of this method approaches the performance of general orthonormal dictionary learning when the number of reflectors increases. Since each reflector is updated individually, this approach is relatively slow due to the large number of matrix manipulations that need to be performed. A natural question is if it possible to decouple the problem such that all reflectors can be updated simultaneously. This idea, which is realized by adding an additional orthogonal constraint of the reflector coefficients, is at the core of the second proposed method. The main benefit of this second approach is that it outperforms the first in terms of running time due to fewer manipulations required, but is slightly inferior in terms of representation quality. Additionally, for this second approach, we are able to perform a detailed performance analysis. While the dictionaries designed by both proposed methods enjoy fast (controllable computational complexity) manipulation the first, slower, approach provides better representation results.

We compare the proposed algorithms in image processing applications, a classical scenario for the evaluation/comparison of sparsifying transforms. We show that the proposed methods cover the full performance range of computational complexity and representation error. Adjusting the number of reflectors in the transform, we can construct anything from dictionaries as fast as the well-known, fixed bases used in image compression with similar representation performance to slower dictionaries that have representation errors matching those of general orthonormal dictionaries. We provide insight into ways of choosing the number of reflectors thus allowing full flexibility to the proposed solutions. Furthermore, we show that in our experimental runs we are always able to construct a fast dictionary that matches the performance of the general orthonormal dictionary with a relative low number of reflectors. Based on these results we conclude that the proposed algorithms are well suited to produce solutions that balance the computational complexity and representation quality of learned dictionaries.

The paper is organized as follows. Section II reviews the concept of orthonormal dictionary learning, Section III presents the proposed algorithms, Section IV provides performance insights into the proposed methods while Section V shows experimentally their effectiveness. 

\section{General orthonormal dictionary learning}
In this section, we review prior work on learning general orthonormal dictionaries and provide some new insights. The objective is to describe the mathematical foundations of the dictionary learning problem, introduce the main notation, formulation and previously proposed solutions.

Given a real dataset $\mathbf{Y} \in \mathbb{R}^{n \times N}$ and sparsity level $s$, the orthonormal dictionary learning algorithm (which we will call Q--DLA) \cite{OrthoDictionary} is formulated as:
\begin{equation}
    \begin{aligned}
        	& \underset{\mathbf{Q},\ \mathbf{X}; \ \mathbf{QQ}^T = \mathbf{Q}^T\mathbf{Q} = \mathbf{I}}{\text{minimize}} & & \|\mathbf{Y}-\mathbf{QX}\|_F^2 \\
			& \text{\ \ \ \ \ \ subject to} &  & \|\mathbf{x}_i\|_{0} \leq s,\ 1 \leq i \leq N,
			\end{aligned}
			\label{eq:dictionaryOrtho}
\end{equation}
where the objective function describes the representation error achieved by the orthonormal dictionary $\mathbf{Q} \in \mathbb{R}^{n \times n}$ with the sparse representations $\mathbf{X} \in \mathbb{R}^{n \times N}$ whose columns are subject to the $\ell_0$ pseudo-norm $\| \mathbf{x}_i \|_0$ (the number of non-zero elements of columns $\mathbf{x}_i$). To avoid trivial solutions, the dimensions obey $s \ll n \ll N$. The problem described in \eqref{eq:dictionaryOrtho} has been extensively studied and used in many applications especially in image processing for compression \cite{SOT, ICIPConf, EMAlgorithm}. Optimizations similar to \eqref{eq:dictionaryOrtho} have been proposed in the past to learn incoherent dictionaries \cite{IncoherentDictionary} or to build initial dictionaries for the general dictionary learning problem \cite{DictInit}.

The solution to \eqref{eq:dictionaryOrtho} proposed in \cite{OrthoDictionary} alternates between computing $\mathbf{X}$ and $\mathbf{Q}$ with one of them fixed, just like in the general dictionary learning case \cite{MOD}. We detail the steps next.

Since the dictionary $\mathbf{Q}$ is orthonormal the sparse representation step reduces to $\mathbf{X} = \mathcal{T}_s(\mathbf{Q}^T\mathbf{Y})$ where $\mathcal{T}_s()$ is an operator that given an input vector zeros all entries except the largest $s$ in magnitude and given an input matrix applies the same operation columnwise. To select the largest entries, per signal, a fast partial sorting algorithm \cite{PartialSort} can be used whose complexity is only $O(n)$.

To solve \eqref{eq:dictionaryOrtho} for variable $\mathbf{Q}$ and fixed $\mathbf{X}$, a problem also known as the orthonormal Procrustes problem \cite{Proc}, a closed form solution $\mathbf{Q} = \mathbf{UV}^T$ is given by the singular value decomposition (SVD) of $\mathbf{YX}^T = \mathbf{U\Sigma V}^T$. Notice that with the representations $\mathbf{X}$ fixed, the reduction in the objective function of \eqref{eq:dictionaryOrtho} achieved by a general orthonormal dictionary $\mathbf{Q}$ is given by:
\begin{equation}
\begin{aligned}
& 	\| \mathbf{Y} - \mathbf{QX} \|_F^2 = \| \mathbf{Y} \|_F^2 + \| \mathbf{X} \|_F^2 + C, \\
&\text{with } C = - 2\text{tr}(\mathbf{Q}^T\mathbf{YX}^T).
\end{aligned}
\label{eq:costDetailed3}
\end{equation}
Develop further to reach
\begin{equation}
\text{tr}(\mathbf{Q}^T\mathbf{YX}^T) = \text{tr}(\mathbf{V}\mathbf{U}^T \mathbf{U} \mathbf{\Sigma V}^T) = \text{tr}(\mathbf{\Sigma}) = \| \mathbf{YX}^T \|_*.
\label{eq:theobjofortho}
\end{equation}
Thus, the reduction in the objective function is $2 \| \mathbf{YX}^T \|_*$. This shows that when considering orthonormal dictionaries, the learning problem can be seen as a nuclear norm maximization with sparsity constraints (and with $\| \mathbf{X} \|_F^2 \leq \| \mathbf{Y} \|_F^2$ to avoid trivial unbounded solutions). Also, notice that at the optimum we have the symmetric positive semidefinite
\begin{equation}
	\mathbf{Q}^T\mathbf{YX}^T = \mathbf{V} \mathbf{\Sigma V}^T, \ \mathbf{QXY}^T = \mathbf{U} \mathbf{\Sigma U}^T.
	\label{eq:symmetry}
\end{equation}
The two are identical since $\mathbf{Q}^T (\mathbf{QXY}^T) \mathbf{Q} = \mathbf{XY}^T\mathbf{Q} = (\mathbf{Q}^T\mathbf{YX}^T)^T = \mathbf{Q}^T\mathbf{YX}^T$, i.e, $\mathbf{V} = \mathbf{Q}^T \mathbf{U}$.

\noindent\textbf{Remark 1.} A positive semidefinite condition for the symmetric $\mathbf{X} (\mathbf{Q}^T\mathbf{Y})^T$, based on the Gershgorin disk theorem, can be stated. Starting from the positive semidefinite condition \eqref{eq:symmetry}, the focus falls on the spectral properties of the symmetric $\mathbf{R} = \mathbf{X} (\mathbf{Q}^T\mathbf{Y})^T = \mathcal{T}_s(\mathbf{Q}^T\mathbf{Y}) (\mathbf{Q}^T\mathbf{Y})^T$. The diagonal elements of this matrix are positive since they are the squared $\ell_2$ norms of the rows of $\mathcal{T}_s(\mathbf{Q}^T\mathbf{Y})$ and moreover they have relative large magnitude since the sparse representation step keeps only the largest $s$ entries $\left(\text{in fact }\text{tr}(\mathbf{R}) = \| \mathcal{T}_s( \mathbf{Q}^T \mathbf{Y}) \|_F^2 = \| \mathbf{X} \|_F^2\right)$. 
Therefore, we can assume that $\mathbf{R}$ is diagonally dominant. We also assume that we eliminate zero rows or rows with very few non-zero entries from $\mathbf{R}$, which corresponds to having atoms in the dictionary that are never/rarely used in the representations. To be more precise, let us denote by $\mathbf{\phi}_i^T$ the $i^\text{th}$ row of $\mathbf{Q}^T\mathbf{Y}$ and with $\mathbf{\psi}_j^T$ the $j^\text{th}$ row of $\mathcal{T}_s(\mathbf{Q}^T\mathbf{Y})$ and we have that $R_{jj} = \mathbf{\psi}_j^T \mathbf{\phi}_j = \mathbf{\psi}_j^T \mathbf{\psi}_j$ and $R_{ij} =  \mathbf{\psi}_i^T \mathbf{\phi}_j$ and, by Gershgorin's disk theorem, the conditions for a positive semidefinite $\mathbf{R}$ are:
\begin{equation}
\mathbf{\psi}_j^T \mathbf{\psi}_j \leq \sum_{i=1, i \neq j}^n | \mathbf{\psi}_j^T \mathbf{\phi}_i| \leq (n-1)\mu,
\label{eq:positivedef}
\end{equation}
for $j = 1,\dots,n$ and where $\mu = \max_{i\neq j} | \mathbf{\psi}_j^T \mathbf{\phi}_i|$. 

The result states that if rows of $\mathbf{Q}^T\mathbf{Y}$ are weakly correlated with the rows of $\mathcal{T}_s(\mathbf{Q}^T\mathbf{Y})$, except for the rows with the same indices, then the pair $(\mathbf{Q}, \mathbf{X})$ is a local minimum of the orthonormal dictionary learning problem.$\hfill \blacksquare$

\noindent \textbf{Remark 2.} Given a dataset $\mathbf{Y}$ and its factorization in a general dictionary $\mathbf{D}$ with sparse representations $\mathbf{X}$, there is no orthonormal transformation $\mathbf{Q}$ such that $\mathbf{QD}$ achieves better representation than $\mathbf{D}$ if $\mathbf{Y}\mathbf{D}^T\mathbf{X}^T$ is symmetric.

\noindent\textit{Proof.} Consider the Procrustes optimization problem in variable $\mathbf{Q}$:
\begin{equation}
	\underset{\mathbf{Q}; \ \mathbf{QQ}^T = \mathbf{Q}^T\mathbf{Q} = \mathbf{I}}{\text{minimize}} \| \mathbf{Y} - \mathbf{QDX} \|_F^2,
\end{equation}
and notice that the minimizer is $\mathbf{Q} = \mathbf{U U}^T = \mathbf{I}$ given that $\mathbf{Y}\mathbf{D}^T\mathbf{X}^T = \mathbf{U\Sigma U}^T$ is symmetric.$\hfill \blacksquare$

\noindent\textbf{Remark 3.} A necessary condition that a general orthonormal dictionary $\mathbf{Q}$ with representations $\mathbf{X}$ is a local minimum of the dictionary learning problem is that $\| \mathbf{X}  \|_F^2 = \| \mathbf{YX}^T \|_*$. For a general overcomplete dictionary $\mathbf{D}$ with representations $\mathbf{X}$, the necessary condition reads $\text{tr}(\mathbf{Y}\mathbf{X}^T\mathbf{D}^T) = \| \mathbf{Y}\mathbf{X}^T\mathbf{D}^T \|_*$.

\noindent\textit{Proof.} With the optimum choice of $\mathbf{Q}$ from the Procrustes result, $\mathbf{QXY}^T$ and $\mathbf{Q}^T\mathbf{YX}^T$ are symmetric by \eqref{eq:symmetry}. By matching the objective function value from \eqref{eq:costDetailed3} with the performance of the orthonormal dictionary from \eqref{eq:theobjofortho}, for fixed $\mathbf{X}$ and $\mathbf{Y}$ there is no general orthonormal dictionary $\mathbf{Q}$ that provides better representation performance than the identity dictionary $\mathbf{I}$ if
\begin{equation}
\text{tr}(\mathbf{YX}^T) = \| \mathbf{YX}^T \|_*,
\label{eq:localoptimumU}
\end{equation}
which holds for example whenever $\mathbf{YX}^T$ is normal (orthogonal, symmetric or skew-symmetric in general) and positive semidefinite -- notice that orthogonal and positive definite just mean that the solution to the Procrustes problem is $\mathbf{Q} = \mathbf{I}$ because $\mathbf{YX}^T = \mathbf{I}$. 
In general, following the same reasoning, we know from \eqref{eq:localoptimumU} that general orthonormal $\mathbf{Q}$ with representations $\mathbf{X} = \mathcal{T}_s(\mathbf{Q}^T\mathbf{Y})$ is a local minimum when
\begin{equation}
\text{tr}(\mathbf{Y}\mathbf{X}^T\mathbf{Q}^T) = \| \mathbf{Y}\mathbf{X}^T\mathbf{Q}^T \|_*.
\label{eq:theequation}
\end{equation}
Finally, using the fact that $\| \mathbf{Y}\mathbf{X}^T\mathbf{Q}^T \|_* =  \| \mathbf{Y}\mathbf{X}^T \|_*$ and that $\text{tr}(\mathbf{Y}\mathbf{X}^T\mathbf{Q}^T) = \text{tr}(\mathbf{Q}^T\mathbf{Y}\mathbf{X}^T) = \| \mathbf{X} \|_F^2$ we reach
\begin{equation}
\| \mathbf{X} \|_F^2 =  \| \mathbf{Y}\mathbf{X}^T \|_*,
\end{equation}
and therefore the objective function in \eqref{eq:costDetailed3} takes the value $\| \mathbf{Y} \|_F^2 - \| \mathbf{X} \|_F^2$.
Equation \eqref{eq:theequation} is also a necessary condition for the local optimality of a general overcomplete dictionary $\mathbf{D}$ with representations $\mathbf{X}$, i.e., $\text{tr}(\mathbf{Y}\mathbf{X}^T\mathbf{D}^T) = \| \mathbf{Y}\mathbf{X}^T\mathbf{D}^T \|_*$ meaning that there is no orthonormal transformation that improves the representation performance of $\mathbf{D}$.$\hfill \blacksquare$

Previous work in the literature deals with the description of local minimum $(\mathbf{D}, \mathbf{X})$ of general dictionary learning schemes \cite{DictionaryIdentification, DictionaryIdentification2}, while other work is concerned with the sample complexity of recovering a dictionary \cite{Spielman12-pp, SampleComplexity, ProvableDictionaryLearning, AlternatingMinimization} under various statistical assumptions and dictionary dimensions. The general analysis in \cite{SampleOfFactorizations} provides sample complexity estimates to control how much the empirical average deviates from the expected objective functions of matrix factorization problems.

As with any alternating minimization solution, the initialization procedure plays an important role. For Q--DLA, our experimental findings show that a very good initial point is the orthonormal basis $\mathbf{Q}$ created from the SVD of the dataset: $\mathbf{Y} = \mathbf{Q \Sigma V}^T$. This choice is also intuitive \cite{DictInit}. A full factorization of $\mathbf{Y}$ is not necessary since we are interested only in the basis $\mathbf{Q}$. As such, a reduced or so called economy size SVD can be performed. Still, depending on the size of the dataset $N$, this step can become expensive in terms of running time. In this paper we propose to approximate $\mathbf{Q}$ with a new orthonormal basis $\mathbf{\bar{Q}}$ obtained by:
\begin{enumerate}
    \item Approximate first $\bar{n} \ll n$ principal components in by using iterative methods \cite{Arnoldi}.
    
    \item Complete the partial structure with random components to obtain the full basis. Finalize by QR orthogonalization to get $\mathbf{\bar{Q}}$.
\end{enumerate}

This initialization works well because typically the lowest singular values of a dataset consisting of real world data have low magnitude.

There are several limitations associated with conventional orthonormal dictionaries. Although the sparse representation step is fast when using an orthonormal dictionary, i.e., no matching \cite{GreedIsGood} or basis pursuit \cite{JustRelax} is necessary and only correlations need to be computed, the representation performance is inferior to that of general dictionaries while the computational complexity is comparable to these dictionaries. For this reason we now move to explore transform structures that allow for a computationally cheaper orthonormal dictionary without destroying the sparsifying properties.

\section{A Householder approach to orthonormal dictionary learning}
In this section, we describe our new approach for dictionary learning based on Householder reflectors. We use the same alternative optimization procedure generally used for dictionary learning and described in Section II. Since we are using orthonormal dictionaries, the sparse approximation step is the same, and thus the focus falls on the dictionary update step which is detailed in this section. 

Therefore, we start by analyzing the properties of Householder reflectors and then introduce two dictionary learning procedures that build orthonormal dictionaries directly factorized into a product of reflectors. We finish the discussion by making some considerations on the initialization of the proposed methods.

\subsection{Householder reflectors for dictionary learning}

Let $\mathbf{u}_1 \in \mathbb{R}^n$ be a normalized vector, i.e., $\| \mathbf{u}_1 \|_2 = 1$. We define the orthonormal symmetric Householder reflector $\mathbf{U}_1 \in \mathbb{R}^{n \times n}$ as
\begin{equation}
	\mathbf{U}_1 = \mathbf{I} - 2\mathbf{u}_1\mathbf{u}_1^T.
	\label{eq:Reflector}
\end{equation}
The reflector $\mathbf{U}_1$ is completely defined by the vector $\mathbf{u}_1$ and as such they may be used equivalently to refer to the reflector. Given a Householder reflector $\mathbf{U}_1 \in \mathbb{R}^{n \times n}$ and a vector $\mathbf{x} \in \mathbb{R}^n$, the product
\begin{equation}
	\mathbf{U}_1\mathbf{x} = \left( \mathbf{I} - 2\mathbf{u}_1\mathbf{u}_1^T \right) \mathbf{x} = \mathbf{x} - 2\mathbf{u}_1 ( \mathbf{u}_1^T\mathbf{x} ) = \mathbf{x} - \nu \mathbf{u}_1,
	\label{eq:ReflectorMultiplication}
\end{equation}
where $\nu = 2 \mathbf{u}_1^T \mathbf{x}$. The computational complexity of \eqref{eq:ReflectorMultiplication} is $N_\text{op} = 4n$, an order of magnitude lower than the general matrix-vector multiplication complexity of $N_\text{op} = n(2n - 1)$. Given $\mathbf{X} \in \mathbb{R}^{n \times N}$ a result similar to \eqref{eq:ReflectorMultiplication} also holds for matrix-matrix multiplication
\begin{equation}
	\mathbf{U}_1\mathbf{X} = \left( \mathbf{I} - 2\mathbf{u}_1\mathbf{u}_1^T \right) \mathbf{X} = \mathbf{X} - \mathbf{u}_1\mathbf{v}_1^T,
\end{equation}
where $\mathbf{v}_1 = 2\mathbf{X}^T\mathbf{u}_1$.

Householder reflectors are often used to introduce zeros in the entries of vectors and to reduce full matrices to upper (or lower) triangular forms with applications to computing least square solutions and QR decompositions. Given a general orthonormal basis $\mathbf{Q} \in \mathbb{R}^{n \times n}$, there exists a sequence of $n-1$ Householder reflectors $\mathbf{U}_j$ such that the following factorization holds:
\begin{equation}
	\mathbf{Q} = \mathbf{U}_{n-1} \mathbf{U}_{n-2} \cdots \mathbf{U}_1 \mathbf{D},
\end{equation}
where $\mathbf{D}$ is a diagonal matrix of size $n \times n$ with entries $D_{ii} = \{ \pm 1 \}, i = 1,\dots,n$. This result follows from the QR factorization of a unitary matrix with Householder reflectors, and from the facts that an orthonormal upper (or lower) triangular matrix is actually diagonal and a product of unitary matrices is itself orthonormal. In this case the reflectors enjoy additional sparse structure since the reflectors vectors $\mathbf{u}_j$ have the first $j-1$ entries set to zero. In the following section we will consider general reflector vectors without any sparsity assumptions. Furthermore we will consider products of $m$ Householder reflectors with $m \ll n$ which will open the way to orthonormal dictionaries that can be manipulated fast. Related work explores the ways of representing an orthonormal basis \cite{OrthoBasis}.

In this section we describe algorithms to learn an orthonormal dictionary $\mathbf{U} \in \mathbb{R}^{n \times n}$ that is a product of a few Householder reflectors, balancing performance and computational complexity. We consider dictionaries with the following structure:
\begin{equation}
\mathbf{U} = \mathbf{U}_m \mathbf{U}_{m-1} \cdots \mathbf{U}_2 \mathbf{U}_1,
\label{eq:MultipleHouseholderDict}
\end{equation}
where all $\mathbf{U}_j$ are Householder reflectors and the number $m$ is on the order $O(\log n)$. Of course, we have that all $\| \mathbf{u}_j \|_2 = 1$. For brevity we do not copy these constraints, but consider them imposed.

\subsection{Learning products of Householder reflectors: an extra orthonormal constraint}

We first explore matrix structures that allow for the simultaneous update of all reflectors in the product $\mathbf{U}$. We keep the same overall dictionary formulation as in \eqref{eq:MultipleHouseholderDict} but with the additional constraint that the reflector vectors obey $\mathbf{u}_i^T \mathbf{u}_j = 0 \text{ for all } i \neq j$.
With this orthogonal constraint the new overall orthonormal symmetric dictionary is
\begin{equation}
\mathbf{U} = \mathbf{U}_m \mathbf{U}_{m-1} \cdots \mathbf{U}_2 \mathbf{U}_1 =  \mathbf{I} - 2 \sum_{j=1}^m \mathbf{u}_j \mathbf{u}_j^T.
\label{eq:MultipleOrthoHouseholderDict}
\end{equation}

Using the fact that the reflector vectors $\mathbf{u}_j$ are orthogonal, the objective function simplifies as
\begin{equation}
\begin{aligned}
\| \mathbf{Y} - \mathbf{UX} \|_F^2 =  & \left\| \mathbf{Y} - \mathbf{X} + 2\sum_{j=1}^m \mathbf{u}_j\mathbf{u}_j^T \mathbf{X} \right\|_F^2 \\
= & \| \mathbf{Y} - \mathbf{X} \|_F^2 + \sum_{j=1}^m \mathbf{u}_j^T \mathbf{Z} \mathbf{u}_j,
\end{aligned}
\label{eq:householdersymmetric}
\end{equation}
where we have defined
\begin{equation}
		\mathbf{Z} = 2(\mathbf{XY}^T + \mathbf{YX}^T) = 2\mathbf{\tilde{Z}}.
	\label{eq:theFirstZ}
\end{equation}
To minimize \eqref{eq:householdersymmetric}, the reflector vectors $\mathbf{u}_j$ are chosen to be the eigenvectors associated with the lowest $m$ negative eigenvalues of $\mathbf{Z}$ (assuming that $m$ negative eigenvalues of $\mathbf{Z}$ exist). Since $\mathbf{Z}$ is symmetric, its eigenvectors are orthonormal and thus obey the constraint that we consider on the reflector vectors $\mathbf{u}_j$. If $\mathbf{Z}$ does not possess $m$ negative eigenvectors then fewer than $m$ reflectors should be constructed, the rest up to $m$ can be set to the zero vector (the reflector becomes the identity).

The full proposed learning procedure, which we call QH$_m$--DLA, is detailed in Algorithm 1. Notice that the product $\mathbf{U}^T \mathbf{Y}$ in the computation of $\mathbf{X}$, step 3) of the iterative process, can be efficiently carried out by using the Householder factorization of $\mathbf{U}$ (complexity $O(nN \log n)$ instead of $O(n^2N)$). This is due to the numerical efficiency of the dictionary $\mathbf{U}$.

The updates of the reflectors in $\mathbf{U}$ and of the sparse representations $\mathbf{X}$ are done exactly at each alternating step of the algorithm and thus the objective function decreases monotonically to a local optimum.

Additionally, QH$_m$--DLA is satisfactory from a theoretically perspective since, as we will see, it allows performance analysis and comparison with Q--DLA. Furthermore, notice that the orthonormal dictionaries created by QH$_m$--DLA are also symmetric.

If we consider a general dictionary $\mathbf{D}$ for sparse representations, then the pair dictionary/representations $(\mathbf{D}, \mathbf{X})$ is equivalent to the pair $(-\mathbf{D}, -\mathbf{X})$ \cite{DictionariesHard2006}. In our setup, notice that if $\mathbf{U}_1$ is a Householder reflector then $-\mathbf{U}_1$ cannot be constructed by \eqref{eq:Reflector}, as $\mathbf{U}_1$ is. Now assume that the matrix $\mathbf{T} = \begin{bmatrix} \mathbf{u}_1 & \mathbf{u}_2 & \dots & \mathbf{u}_n \end{bmatrix}$ contains all $n$ eigenvectors of $\mathbf{\tilde{Z}}$ ordered in increased order of their corresponding eigenvalues. Let $\mathbf{T}_{i:j}$ denote a matrix consisting of all the reflector vectors from the $i^\text{th}$ to the $j^\text{th}$ column of $\mathbf{T}$. Then, due to $\mathbf{T}_{1:i}\mathbf{T}_{1:i}^T + \mathbf{T}_{i+1:n}\mathbf{T}_{i+1:n}^T = \mathbf{I}$, we have that:
\begin{equation}
	-(\mathbf{I} - 2\mathbf{T}_{1:i}\mathbf{T}_{1:i}^T) = \mathbf{I} - 2\mathbf{T}_{i+1:n}\mathbf{T}_{i+1:n}^T.
\end{equation}
This shows that there is a correspondence in performance according to the number of reflectors that are selected: with the first $m$ reflectors we have the dictionary $\mathbf{U}$ (and representations $\mathbf{X}$) while with the other $n-m$ reflectors we have the dictionary $-\mathbf{U}$ (and representations $-\mathbf{X}$).
\begin{algorithm}[t]
	\caption{ \textbf{-- QH$_m$--DLA (Orthogonal Householder Dictionary Learning Algorithm).} \newline \textbf{Input: } The dataset $\mathbf{Y} \in \mathbb{R}^{n \times N}$, the number of Householder reflectors in the transform $m$, the target sparsity $s$ and the maximum number of iterations $K$. \newline \textbf{Output: } The sparsifying transform $\mathbf{U} = \mathbf{U}_m \cdots \mathbf{U}_1$ with $\mathbf{u}_i^T \mathbf{u}_j = 0,\ i \neq j$ and sparse representations $\mathbf{X}$ such that $\| \mathbf{Y} - \mathbf{UX} \|_F^2$ is reduced.}
	\begin{algorithmic}
		\State \textbf{Initialization:}
		\begin{enumerate}
			\setlength{\itemindent}{+.25in}
			\item Perform the economy size singular value decomposition of size $m+1$ of the dataset $\mathbf{Y} = \mathbf{Q} \mathbf{\Sigma} \mathbf{V}^T$.
			\item Reduce $\mathbf{Q} \in \mathbb{R}^{n \times (m+1)}$ to an upper triangular form with Householder reflectors defined by $\mathbf{u}_1,\ldots,\mathbf{u}_m$. The reflector that introduces zeros in the first column is $\mathbf{u}_m$.
			\item Orthogonalize $\mathbf{u}_1,\dots,\mathbf{u}_m$ by the QR algorithm.
			\item Compute sparse representations $\mathbf{X} = \mathcal{T}_s(\mathbf{U}^T \mathbf{Y})$.
		\end{enumerate}
		
		\State \textbf{Iterations} $1,\dots,K$:
		\begin{enumerate}
			\setlength{\itemindent}{+.25in}
			\item Construct the matrix: $\mathbf{\tilde{Z}} = \mathbf{XY}^T + \mathbf{YX}^T$.
			\item Compute the $m$ lowest eigenvalue/eigenvector pairs of $\mathbf{\tilde{Z}}$. Set to $\mathbf{0}$ the eigenvectors associated to nonnegative eigenvalues. Update reflector vectors $\mathbf{u}_j$ with the eigenvectors just computed. The eigenvector of the lowest negative eigenvalue goes to $\mathbf{u}_m$.
			\item Compute sparse representations $\mathbf{X} = \mathcal{T}_s(\mathbf{U}^T \mathbf{Y})$.
		\end{enumerate}
	\end{algorithmic}
\end{algorithm}

\subsection{Learning products of Householder reflectors: the unconstrained case}

We again consider the case where the dictionary $\mathbf{U}$ has the structure from \eqref{eq:MultipleHouseholderDict} but now no additional constraints are assumed on the reflectors. This time we update each reflector sequentially. The new objective function becomes $\| \mathbf{Y} - \mathbf{U}_m \mathbf{U}_{m-1} \cdots \mathbf{U}_2 \mathbf{U}_1 \mathbf{X} \|_F^2$. To optimize the $j^\text{th}$ Householder reflector, we write the objective function as
\begin{equation}
	\| \left(  \mathbf{U}_{j+1} \cdots \mathbf{U}_m  \right)  \mathbf{Y} - \mathbf{U}_j \left( \mathbf{U}_{j-1} \cdots \mathbf{U}_1 \right)  \mathbf{X} \|_F^2,
	\label{eq:UpdateOnlyOne}
\end{equation}
where we have used that all unitary matrices, and thus Householder reflectors, preserve the Frobenius norm and the fact that the reflectors are symmetric:
\begin{equation}
	\| \mathbf{Y} - \mathbf{U}_1\mathbf{X}  \|_F^2 = \| \mathbf{U}_1^T\mathbf{Y} - \mathbf{X} \|_F^2 = \|\mathbf{U}_1\mathbf{Y} - \mathbf{X} \|_F^2.
\end{equation}
We have now reduced the problem to the QH$_1$--DLA case for the updated dataset $\left(  \mathbf{U}_{j+1} \cdots \mathbf{U}_m  \right)  \mathbf{Y}$ and the updated representations $\left( \mathbf{U}_{j-1} \cdots \mathbf{U}_1 \right)  \mathbf{X}$. Following the same computation that leads to \eqref{eq:theFirstZ}, we now reach that the best update for the fixed $\mathbf{u}_j$ is the eigenvector associated with the lowest negative eigenvalue of
\begin{equation}
	\begin{aligned}
		\mathbf{Z} = & 2 \left( \mathbf{U}_{j-1} \cdots \mathbf{U}_1 \right) \mathbf{XY}^T \left(  \mathbf{U}_{j+1} \cdots \mathbf{U}_m  \right)^T  + \\
			 & \quad \quad \quad 2 \left(  \mathbf{U}_{j+1} \cdots \mathbf{U}_m  \right) \mathbf{YX}^T   \left( \mathbf{U}_{j-1} \cdots \mathbf{U}_1 \right)^T.
	\end{aligned}
	\label{eq:theSecondZ}
\end{equation}

Each reflector in the product of $\mathbf{U}$ is updated sequentially in this manner. The full procedure, which we call H$_m$--DLA, is detailed in Algorithm 2. We expect the performance of this algorithm to be in general inferior to that of Q--DLA in terms of representation error, approaching it as $m$ approaches $n$, and to be superior to that of QH$_m$--DLA, due to the missing additional orthogonal constraints. Still, since all reflectors are computed together and no extensive matrix manipulation is required QH$_m$--DLA runs faster than H$_m$--DLA. This opens the possibility of using QH$_m$--DLA as an initialization procedure for H$_m$--DLA.  Finally, QH$_1$--DLA and H$_1$--DLA are equivalent. Also notice that the computation of $\mathbf{\tilde{Z}}$ can be optimized across the iterative process in step 1a): denote $\mathbf{R}_j = \left( \mathbf{U}_{j-1} \cdots \mathbf{U}_1 \right) \mathbf{XY}^T \left(  \mathbf{U}_m \cdots \mathbf{U}_{j+1} \right)$ from the $j^\text{th}$ iteration, the for the next iteration when computing $\mathbf{U}_{j+1}$ we simply have that $\mathbf{R}_{j+1}=\mathbf{U}_j \mathbf{R}_j \mathbf{U}_{j+1}^T$ -- which can be done efficiently by left and right reflector multiplication formulas.

Just as in the case of QH$_m$--DLA, the update of each reflector $\mathbf{U}_j$ and of the representations $\mathbf{X}$ are done by solving exactly the optimization problems (with the other variables fixed) and thus the objective function monotonically decreases to a local minimum point.
\begin{algorithm}[t]
\caption{ \textbf{-- H$_m$--DLA (Householder Dictionary Learning Algorithm).} \newline \textbf{Input: } The dataset $\mathbf{Y} \in \mathbb{R}^{n \times N}$, the number of Householder reflectors in the transform $m$, the target sparsity $s$ and the maximum number of iterations $K$. \newline \textbf{Output: } The sparsifying transform $\mathbf{U} = \mathbf{U}_m \cdots \mathbf{U}_1$ and sparse representations $\mathbf{X}$ such that $\| \mathbf{Y} - \mathbf{UX} \|_F^2$ is reduced.}
\begin{algorithmic}
	\State \textbf{Initialization:}
		\begin{enumerate}
			\setlength{\itemindent}{+.25in}
			\item Perform the economy size singular value decomposition of size $m+1$ of the dataset $\mathbf{Y} = \mathbf{Q}\mathbf{\Sigma} \mathbf{V}^T$.
			\item Reduce $\mathbf{Q} \in \mathbb{R}^{n \times (m+1)}$ to an upper triangular form by Householder reflectors defined by $\mathbf{u}_1,\ldots,\mathbf{u}_m$. The reflector that introduces zeros in the first column is $\mathbf{u}_m$.
			\item Compute sparse representations $\mathbf{X} = \mathcal{T}_s(\mathbf{U}^T \mathbf{Y})$.
		\end{enumerate}

	\State \textbf{Iterations} $1,\dots,K$:
		\begin{enumerate}
			\setlength{\itemindent}{+.25in}
			\item For $j = 1,\dots,m$:
				\begin{enumerate}
				\setlength{\itemindent}{+.25in}
					\item Construct the matrix:

\quad \quad $ \mathbf{\tilde{Z}} = \left(  \mathbf{U}_{j-1} \cdots \mathbf{U}_1 \right) \mathbf{XY}^T \left(  \mathbf{U}_{j+1} \cdots \mathbf{U}_{m} \right)^T $,

\quad \quad $\mathbf{\tilde{Z}} = \mathbf{\tilde{Z}} + \mathbf{\tilde{Z}}^T$.

					\item Compute lowest eigenvalue $\lambda_\text{min}$ of $\mathbf{\tilde{Z}}$ with eigenvector $\mathbf{v}$. If $\lambda_\text{min} \geq 0$ then set $\mathbf{v} = \mathbf{0}$. Update reflector vector $\mathbf{u}_j = \mathbf{v}$.
				\end{enumerate}

			\item Compute sparse representations $\mathbf{X} = \mathcal{T}_s(\mathbf{U}^T \mathbf{Y})$.
		\end{enumerate}
\end{algorithmic}
\end{algorithm}

\subsection{The initializations of H$_m$--DLA and QH$_m$--DLA}
Initialization is important for any alternating minimization algorithm. In principle, the proposed methods can be initialized with random reflectors $\mathbf{u}_j$ but the idea is to provide an initialization such that the methods converge in few iterations. The computational complexity of the initialization should be much lower than that of the learning algorithms.

In both the cases of H$_m$--DLA and QH$_m$--DLA, the initialization procedures start by computing the reduced singular value decomposition of size $m$ of the dataset $\mathbf{Y} = \mathbf{Q\Sigma V}^T$. Then $\mathbf{Q}$ is diagonalized by Householder reflectors thus providing the $n$ reflectors. Among these we choose $m$ reflectors to initialize our algorithms. In the case of QH$_m$--DLA the reflectors previously obtained are further orthogonalized by the QR algorithm thus ensuring compliance with all the constraints of the method.

\section{Comments on the proposed algorithms and connections to previous work}

Now that the main algorithms have been described, in this section we examine the achievable representation performance of Householder based dictionaries. 
First, we analyze the simple case of a single Householder reflector dictionary (analysis that is pertinent also to each step of the H$_m$--DLA) and then consider the QH$_m$--DLA. Finally, we show the similarities between the representation error achievable by our proposed dictionaries and that of general orthonormal dictionaries.

\subsection{Performance of a single Householder reflector dictionary}

Considering a dictionary composed of a single Householder reflector, the objective function in \eqref{eq:householdersymmetric} reduces to
\begin{equation}
	\begin{aligned}
	& \| \mathbf{Y} - \mathbf{U}_1 \mathbf{X} \|_F^2  =  \| \mathbf{Y} \|_F^2 + \| \mathbf{X} \|_F^2  + C, \\
	&\text{with } C = - 2\text{tr}(\mathbf{XY}^T) + 2\mathbf{u}_1^T (\mathbf{XY}^T + \mathbf{YX}^T)\mathbf{u}_1.
	\end{aligned}
	\label{eq:costDetailed}
\end{equation}
Assuming some normalization of the dataset like mean subtraction and $\ell_2$ normalization of the columns, it is reasonable to consider $\| \mathbf{Y} \|_F^2 = N$. The norm $ \| \mathbf{X} \|_F^2 $ is maximized in the sparse reconstruction step, where we keep the largest absolute value entries in the representations. The goal is twofold:
\begin{itemize}
	\item Maximize the trace of $\mathbf{XY}^T$.

	\item Minimize the lowest eigenvalue of $\mathbf{\tilde{Z}} = \mathbf{XY}^T + \mathbf{YX}^T$.
\end{itemize}

The two goals are related since $\text{tr}(\mathbf{\tilde{Z}}) = 2\text{tr}(\mathbf{XY}^T)$. Therefore, the performance of our algorithms depends on the spectral properties of $\mathbf{\tilde{Z}}$. In an ideal situation, the lowest, negative, eigenvalue of this matrix should be maximally reduced while the rest of the eigenvalues remain positive and their sum is maximized. 
An ideal case would be that the spectrum obeys $\Lambda(\mathbf{\tilde{Z}}) = \{ -\alpha_1, \beta_1, \ldots, \beta_{n-1} \}$, one negative eigenvalue and $n-1$ non-negative. Now the cost in \eqref{eq:costDetailed} is maximally reduced by the sum of the singular values of $\mathbf{\tilde{Z}}$ also known as its nuclear norm, i.e., $C = - \| \mathbf{\tilde{Z}} \|_* =  -\left(\alpha_1 + \sum_{i=1}^{n-1} \beta_i\right)$.

\subsection{Performance of Householder based dictionaries}

We now analyze the dictionaries created by QH$_m$--DLA. In the case of H$_m$--DLA, since the reflectors are updated sequentially, we defer to the discussion for QH$_1$--DLA.

The case that can be more easily approached from an analysis perspective is the one of QH$_m$--DLA, where all reflectors are updated simultaneously. In this case, the objective function \eqref{eq:householdersymmetric} reduces to
\begin{equation}
	\begin{aligned}
	& \| \mathbf{Y} - \mathbf{U} \mathbf{X} \|_F^2  =  \| \mathbf{Y} \|_F^2 + \| \mathbf{X} \|_F^2  + C, \\
	&\text{with } C = - 2\text{tr}(\mathbf{XY}^T) +  2 \sum_{j=1}^m \mathbf{u}_j^T (\mathbf{XY}^T + \mathbf{YX}^T) \mathbf{u}_j.
	\end{aligned}
	\label{eq:costDetailed2}
\end{equation}

Similar to the single Householder reflector case, the performance depends on the spectrum $\Lambda(\mathbf{\tilde{Z}}) = \{ -\alpha_1, \ldots, -\alpha_m, \beta_1, \dots, \beta_{n-m} \}$. To minimize the objective function in \eqref{eq:costDetailed2}, we need to choose $m$ Householder reflectors corresponding to the $m$ negative eigenvalues in $\Lambda(\mathbf{\tilde{Z}})$. In this way, \eqref{eq:costDetailed2} is maximally reduced by the nuclear norm of $\mathbf{\tilde{Z}}$, i.e., $C = -\|\mathbf{\tilde{Z}}  \|_*=-\left(\sum_{i=1}^m \alpha_i + \sum_{i=1}^{n-m} \beta_i \right)$.

If the spectrum of $\mathbf{\tilde{Z}}$ is non-negative, then no reflector can decrease the objective function and the dictionary is set to $\mathbf{U} = \mathbf{I}$; with the given $\mathbf{Y}$ and $\mathbf{X}$ there is no Householder reflector that can improve upon the representation performance. Equally bad, if the spectrum is non-positive then all $n$ eigenvectors are selected and by \eqref{eq:MultipleOrthoHouseholderDict} it follows that the dictionary is $\mathbf{U} = -\mathbf{I}$. In practice, depending on the magnitude of all the $m$ negative eigenvalues of $\mathbf{\tilde{Z}}$ we may choose a smaller number of reflectors to construct $\mathbf{U}$. Of course, the representation performance is slightly inferior this way but the benefit is a faster transform. The trade-off can be balanced based on application specific requirements.

A situation that is of interest is when the sparse factorization can be done exactly, i.e., $\mathbf{Y} = \mathbf{UX}$. Considering that some normalization has taken place for the dataset such that $\| \mathbf{Y} \|_F^2 = N$ and because orthonormal transformations preserve $\ell_2$ norms we have that $\| \mathbf{X} \|_F^2 = N$, i.e., we have in effect exactly $\mathbf{X} = \mathbf{U}^T\mathbf{Y}$. The objective function of the optimization problem reaches zero and thus the nuclear norm of $\mathbf{\tilde{Z}}$ is maximized to $2N$.

A last comment concerns the addition to the reflector $\mathbf{u}_i$ of the sparse structure typical of QR decompositions, i.e., consider $\mathbf{u}_i = \begin{bmatrix}  \mathbf{0}; & \mathbf{\tilde{u}}_i \end{bmatrix}$. With this new structure the minimizer $\mathbf{\tilde{u}}_i$ of the expression in \eqref{eq:costDetailed} is given by the eigenvector associated with the smallest, negative, eigenvalue of the lower right-hand side square sub-matrix of size $(n-i+1)$ from $\mathbf{\tilde{Z}}$. This structure appears during the initialization step discussed in Section III. 

\subsection{Connections between Householder based dictionaries and general orthonormal dictionaries}

The proposed algorithms are closely connected to the task of learning a general orthonormal dictionary. Increasing $m$ for H$_m$--DLA and QH$_m$--DLA will reduce the performance gap between dictionaries designed by these methods and the orthonormal dictionaries designed via Q--DLA, of course at the cost of higher computational demand. We discuss now some properties and connections between the various dictionary learning procedures.

\noindent\textbf{Remark 4.} Given a dataset $\mathbf{Y}$ represented in the general dictionary $\mathbf{D}$ with the sparse representations $\mathbf{X}$, there is no reflector $\mathbf{U}_1$ such that $\mathbf{U}_1\mathbf{D}$ achieves lower representation error than $\mathbf{D}$ if $\mathbf{DXY}^T + \mathbf{Y}(\mathbf{DX})^T$ is positive semidefinite.

\noindent\textit{Proof.} Check for the existence of a reflector $\mathbf{U}_1$ such that no left dictionary update improves the representation
\begin{equation}
\| \mathbf{Y} - \mathbf{D} \mathbf{X}\|_F^2 > \| \mathbf{Y} - \mathbf{U}_1\mathbf{D}\mathbf{X}\|_F^2.
\label{eq:BetterThanI2}
\end{equation}

If such a reflector does not exist then $\mathbf{D}$ may be viewed as a local minimum (this is a necessary condition). This is equivalent to considering an updated dictionary $\mathbf{U}_1\mathbf{D}$. Therefore, if the symmetric matrix
\begin{equation}
\mathbf{Z}_1 = \mathbf{DXY}^T + \mathbf{Y}(\mathbf{DX})^T,
\label{eq:Cond1}
\end{equation}
is positive semidefinite then $\mathbf{D}$ is a local minimum of, i.e., there is no reflector $\mathbf{U}_1$ such that $\mathbf{U}_1\mathbf{D}$ is able to achieve a lower objective function value in than $\mathbf{D}$. Compare this with Remark 2. $\hfill \blacksquare$



As we have seen in the previous sections, the positive semidefinite condition is necessary and sufficient when describing local minima of the Householder based dictionaries. In the case of general orthonormal and, due to \eqref{eq:BetterThanI2} and \eqref{eq:Cond1}, also general (even overcomplete) dictionaries the condition is necessary, but not sufficient.

\noindent\textbf{Remark 5.} Q--DLA always performs better than QH$_m$--DLA, the performance matches when $\mathbf{YX}^T$ is symmetric. 
We have shown by \eqref{eq:theobjofortho} that the objective function reduction possible by a general orthogonal dictionary is $2\| \mathbf{YX}^T \|_*$. Due to the triangle inequality which is obeyed by the nuclear norm, this quantity is larger or equal at worse to the reduction achievable when using a symmetric dictionary designed via QH$_m$--DLA, which is $\| \mathbf{XY}^T + \mathbf{YX}^T \|_*$. As expected, due to its additional constraints, QH$_m$--DLA performs worse than the Q--DLA. In general, only H$_m$--DLA, with a sufficiently large $m$, has the capability to match the Q--DLA.$\hfill \blacksquare$ 

\noindent\textbf{A simple example in $\mathbb{R}^2$.} To illustrate the previous results, consider a dataset $\mathbf{Y} \in \mathbb{R}^{2 \times N}$ and the initial dictionary $\mathbf{Q} = \mathbf{I}$. With target sparsity $s = 1$ we have, under a permutation of columns to highlight the row structure, the representations
\begin{equation}
	\mathbf{X} = \begin{bmatrix}
					\mathbf{y}_{11}^T & \mathbf{0}^T \\
					\mathbf{0}^T & \mathbf{y}_{22}^T 
				\end{bmatrix} \text{ where } \mathbf{Y} = \begin{bmatrix}
														\mathbf{y}_{11}^T & \mathbf{y}_{12}^T \\
														\mathbf{y}_{21}^T & \mathbf{y}_{22}^T
													\end{bmatrix},
\end{equation}
and therefore
\begin{equation}
	\mathbf{YX}^T = \begin{bmatrix}
					\|\mathbf{y}_{11}\|_2^2 & \mathbf{y}_{12}^T\mathbf{y}_{22} \\
					\mathbf{y}_{11}^T \mathbf{y}_{21} & \|\mathbf{y}_{22}\|_2^2
					\end{bmatrix},\ \mathbf{\tilde{Z}} = \mathbf{YX}^T + \mathbf{XY}^T.
	\label{eq:theYXT}
\end{equation}
By \eqref{eq:positivedef} and with \eqref{eq:theYXT} we see that there is no Householder based dictionary that improves the presentation error if $2 \| \mathbf{y}_{11} \|_2^2 < \mathbf{y}_{11}^T\mathbf{y}_{21} + \mathbf{y}_{22}^T\mathbf{y}_{12}$ and $2 \| \mathbf{y}_{22} \|_2^2 < \mathbf{y}_{11}^T\mathbf{y}_{21} + \mathbf{y}_{22}^T\mathbf{y}_{12}$, i.e., $\mathbf{\tilde{Z}}$ is positive semidefinite. Since $\text{tr}(\mathbf{\tilde{Z}}) = 2\| \mathbf{X} \|_F^2 > 0$ one of the eigenvalues is necessarily positive and therefore the previous two conditions lead to $2\| \mathbf{y}_{11}  \|_2 \| \mathbf{y}_{22} \|_2 < \mathbf{y}_{11}^T\mathbf{y}_{21} + \mathbf{y}_{22}^T\mathbf{y}_{12}$. Therefore, with a Householder based dictionary the possible reduction in the representation error is $\|\mathbf{\tilde{Z}} \|_* = 2\sqrt{ \text{tr}(\mathbf{\tilde{Z}})^2/4 - \det(\mathbf{\tilde{Z}}) } = 2\sqrt{ \|\mathbf{X}\|_F^4 - \det(\mathbf{\tilde{Z}}) }$. The Frobenius norm of the representations is maximized in the sparse approximation step while $-\det(\mathbf{\tilde{Z}})$, which is always positive, is increased when maximizing $\mathbf{y}_{11}^T\mathbf{y}_{21} + \mathbf{y}_{22}^T\mathbf{y}_{12}$.

Assuming $\mathbf{YX}^T$ is positive semidefinite then we know there is no Householder based dictionary that can improve the representations. If we consider now general orthonormal dictionaries with \eqref{eq:theYXT} we know from \eqref{eq:localoptimumU} that if $\mathbf{y}_{12}^T\mathbf{y}_{22} \approx \pm \mathbf{y}_{11}^T \mathbf{y}_{21}$ (i.e., $\mathbf{YX}^T$ is approximately symmetric or skew-symmetric) there is also no orthonormal dictionary that can perform much better in terms of representation than the identity.
$\hfill \blacksquare$

\subsection{Householder reflectors vs. Givens rotations for learning fast dictionaries}

Householder reflectors are not the only elementary building blocks for orthonormal structures. Any orthonormal dictionary of size $n \times n$ can also be factorized in a product of Givens rotations \cite{Golub1996} parameterized by $c, s$ and the indices $(i,j)$ like
	$\mathbf{G}_{ij} = \begin{bmatrix} 
							\mathbf{I}_{i-1} &  &  &  & \\
							 & c &  & s & \\
							 & & \mathbf{I}_{j-i-1} & & \\
							 & -s & & c & \\
							 & & & & \mathbf{I}_{n-j} \\
						\end{bmatrix},\ c^2 + s^2 = 1$.
Givens rotations have been previously used with great success in several matrix factorization applications \cite{Treelets2008, SparseMatrixTransform2011, MultiresolutionMatrixFactorization2014}.

Consider using a single Givens rotation as a dictionary. We reach the optimization problem
		$\underset{c, s, (i,j);\ c^2 + s^2 = 1}{\text{minimize}} \ \  \| \mathbf{Y} - \mathbf{G}_{ij} \mathbf{X} \|_F^2$,
which is equivalent to
		\begin{equation*}
		\underset{c, s, (i,j);\ c^2 + s^2 = 1}{\text{minimize}} 
		\left\| \begin{bmatrix} \mathbf{y}_i^T \\ \mathbf{y}_j^T \end{bmatrix} -
			\begin{bmatrix}
				c & s \\ -s & c
			\end{bmatrix}
				\begin{bmatrix} \mathbf{x}_i^T \\ \mathbf{x}_j^T \end{bmatrix} \right\|_F^2,
		\label{eq:GivensLearning2}
		\end{equation*}
where $\mathbf{y}_i^T$ and $\mathbf{x}_i^T$ are the $i^\text{th}$ rows of $\mathbf{Y}$ and $\mathbf{X}$, respectively.

When indices $(i,j)$ are fixed, the optimization reduces to a two dimensional orthogonal Procrustes problem. While to select the indices $(i,j)$, among the ${n \choose 2}$ possibilities, an appropriate strategy needs to be defined. Indeed, Givens rotations also seem an appropriate tool to approach the fast dictionary learning problem, but it is beyond the scope of this paper to analyze it in detail.

\section{Results}

\begin{figure}[t]
	\centering
	\includegraphics[width=0.38\textwidth]{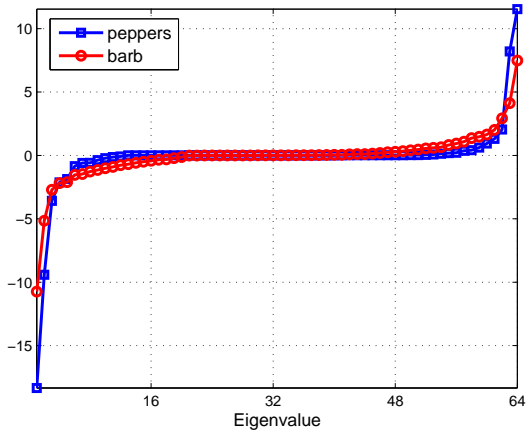}
	\caption{Normalized eigenvalues of $\mathbf{\tilde{Z}}$ after convergence of QH$_m$--DLA for images peppers and barb with sparsity $s = 4$ and  $m = 12$ reflectors.}
	\label{fig:LenaVsBarb}
\end{figure}

\begin{figure}[t]
	\centering
	\includegraphics[width=0.4\textwidth]{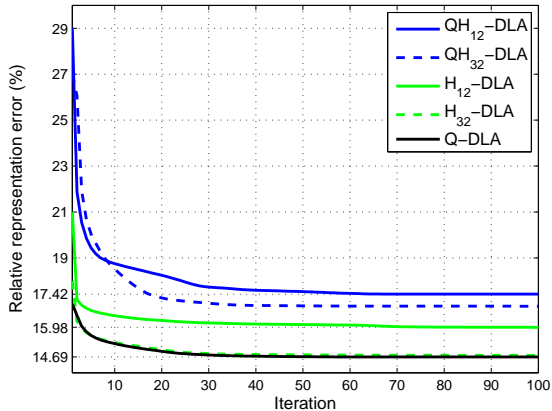}
	\caption{For the proposed methods we show the evolution of the relative representation error $\| \mathbf{Y} - \mathbf{DX} \|_F^2 \| \mathbf{Y} \|_F^{-2}$ for the dataset $\mathbf{Y}$ created from the patches of the images couple, peppers and boat with sparsity $s = 4$ and for $m \in \{ 12, 32\}$ reflectors. For reference we show Q--DLA \cite{OrthoDictionary}.}
	\label{fig:IterationEvolution}
\end{figure}
\def\arraystretch{1.2}
\begin{table*}[t]
	\begin{center}
		\caption{RMSE in the case of several dictionaries computed from known test images. Sparsity level is $s=4$ and the dataset is $\mathbf{Y} \in \mathbb{R}^{64 \times 4096}$ in each case. The learning procedures run after mean extraction and normalization $\mathbf{Y} = \mathbf{Y}/255.$ The best results of the fast dictionaries are shown in bold font.}\label{tb:Table1}
		\begin{tabular}{|c|c<{\hspace{-3pt}}|c<{\hspace{-3pt}}|c<{\hspace{-3pt}}|c<{\hspace{-3pt}}|c<{\hspace{-3pt}}|c<{\hspace{-3pt}}|c<{\hspace{-3pt}}|c<{\hspace{-3pt}}|c<{\hspace{-3pt}}|c<{\hspace{-3pt}}|c<{\hspace{-3pt}}|c<{\hspace{-3pt}}|}
			\hline
			& peppers & boat & pollen & mri & cameraman & pirate & barb & baboon & hill & couple & house & fingerprint \\ \hline
			DCT & 0.0395 &0.0419 & 0.0461 & 0.0721 & 0.0619 & 0.0507 & \textbf{0.0435} & 0.0694 & 0.0361 & 0.0432 & 0.0374 & 0.0765 \\ \hline
			H$_6$--DLA & 0.0294 & 0.0371 & 0.0421 & 0.0649  & 0.0568 & 0.0453 & 0.0508 & 0.0738 & 0.0331 & 0.0405 & 0.0298 & 0.0536 \\ \hline
			H$_{12}$--DLA & \textbf{0.0261} & \textbf{0.0324}  & \textbf{0.0376} & \textbf{0.0611} & \textbf{0.0512} & \textbf{0.0421} & 0.0436 & \textbf{0.0691} & \textbf{0.0302}& \textbf{0.0353} & \textbf{0.0255} & \textbf{0.0497}  \\ \hline
			QH$_6$--DLA & 0.0306 & 0.0375 & 0.0425 & 0.0656  & 0.0575  & 0.0457 & 0.0508  & 0.0739 & 0.0334 & 0.0411& 0.0302 & 0.0542  \\ \hline
			QH$_{12}$--DLA & 0.0278 & 0.0336 & 0.0388 & 0.0626 & 0.0533  & 0.0434 & 0.0444 & 0.0702 & 0.0313 & 0.0366 & 0.0275 & 0.0512 \\ \hline \hline
			H$_{32}$--DLA & 0.0253 & 0.0310 & 0.0371 & 0.0594 & 0.0472 & 0.0407  & 0.0348 & 0.0649 & 0.0288 & 0.0336 & 0.0234 & 0.0492 \\ \hline
			QH$_{32}$--DLA & 0.0278 & 0.0332 & 0.0385 & 0.0617 & 0.0519 & 0.0428 & 0.0397 & 0.0681 & 0.0305 & 0.0364 & 0.0265 & 0.0511 \\ \hline
			Q--DLA \cite{OrthoDictionary} & 0.0256 & 0.0312 & 0.0372  & 0.0596 & 0.0473 & 0.0409 & 0.0361 & 0.0654 & 0.0292 & 0.0339 & 0.0241 & 0.0496 \\ \hline
			SK--SVD \cite{SKSVD} & 0.0191 & 0.0231 & 0.0275 & 0.0462 & 0.0311 & 0.0328 & 0.0266 & 0.0561 & 0.0235 & 0.0266 & 0.0143 & 0.0344 \\ \hline
		\end{tabular}
	\end{center}
\end{table*}

\begin{figure}[t]
	\centering
	\includegraphics[width=0.4\textwidth]{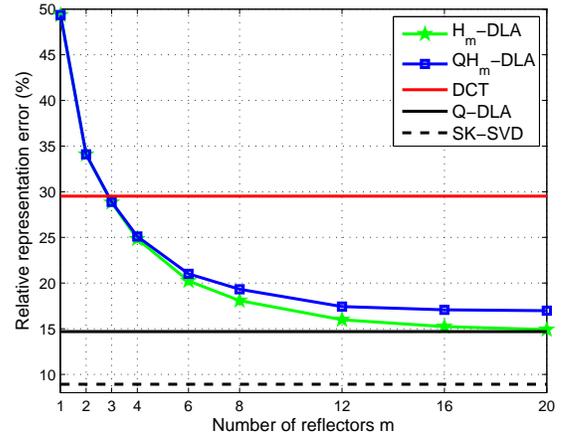}
	\caption{Relative representation error $\| \mathbf{Y} - \mathbf{DX} \|_F^2 \| \mathbf{Y} \|_F^{-2}$, in percent, for the proposed algorithms with the dataset composed of all patches from the images couple, peppers and boat for sparsity $s=4$. For reference we show the DCT, Q--DLA \cite{OrthoDictionary} and SK--SVD \cite{SKSVD}.}
	\label{fig:HouseholderComparisons}
\end{figure}
\begin{figure}[t]
	\centering
	\includegraphics[width=0.38\textwidth]{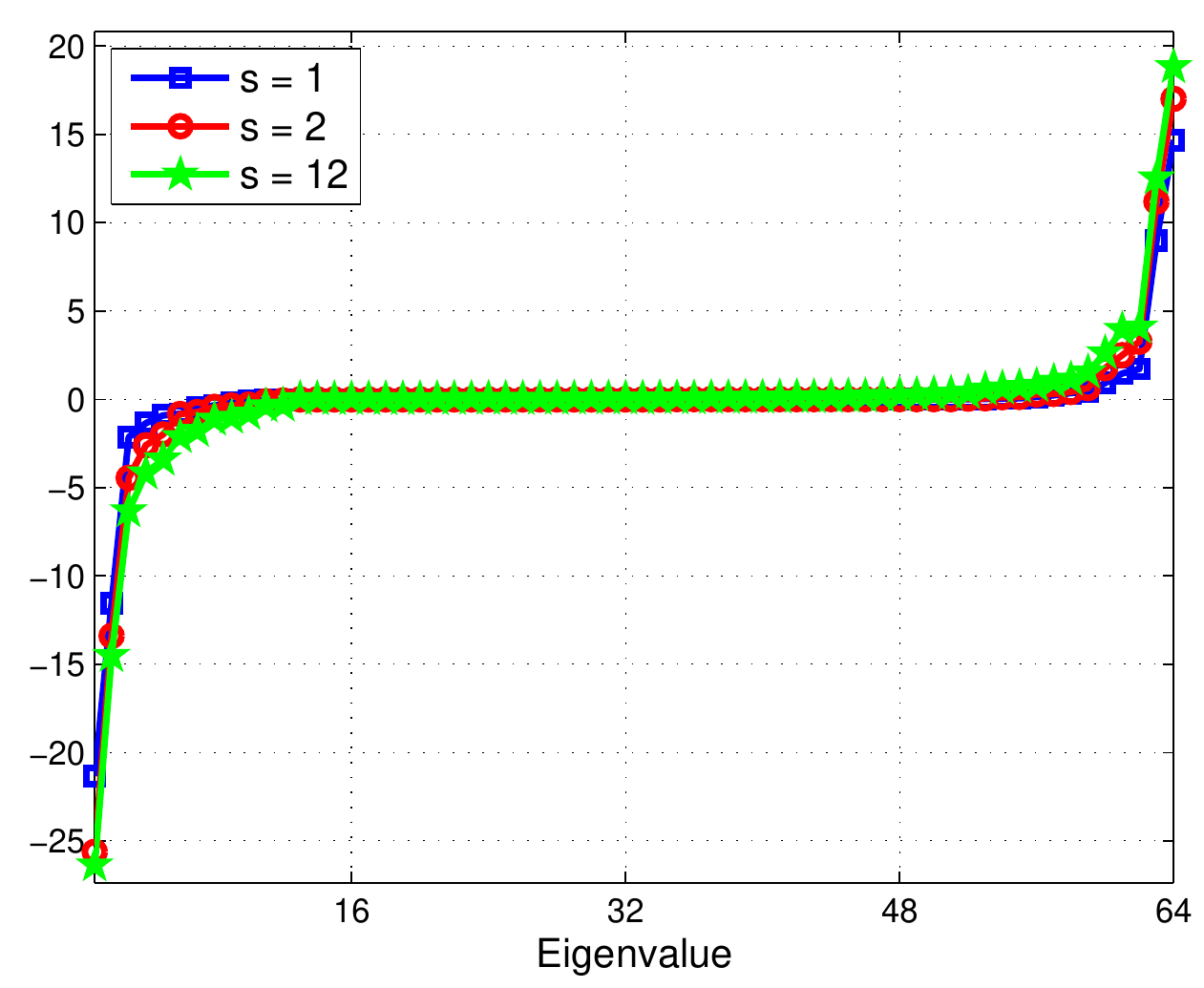}
	\caption{Normalized eigenvalues of $\mathbf{\tilde{Z}}$ after convergence of QH$_{12}$--DLA with various sparsity levels for the dataset in Figure \ref{fig:HouseholderComparisons}.}
	\label{fig:Plot_of_sparsity}
\end{figure}

\begin{figure}[t]
	\centering
	\includegraphics[width=0.38\textwidth]{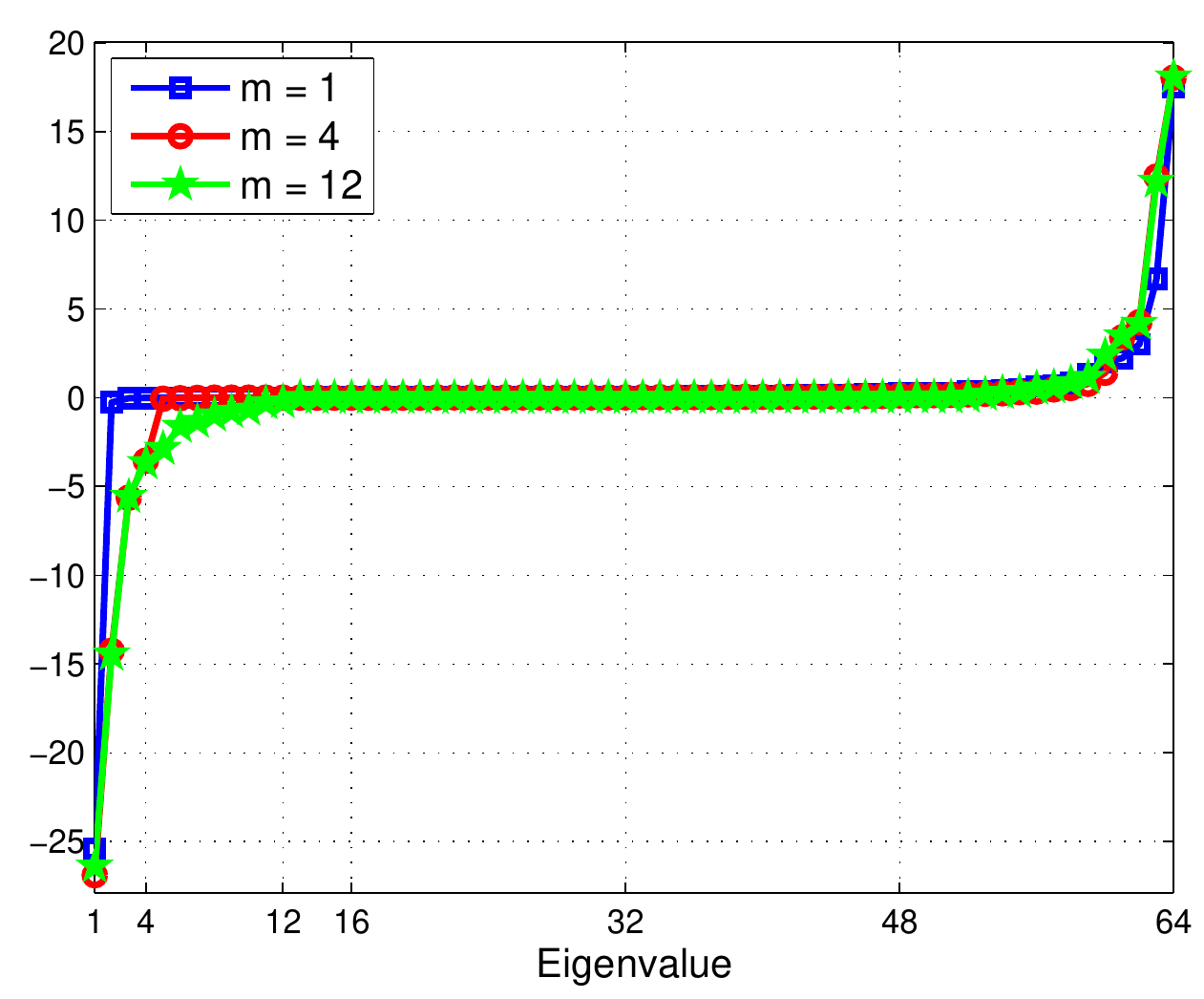}
	\caption{Normalized eigenvalues of $\mathbf{\tilde{Z}}$ after convergence of QH$_m$--DLA with sparsity $s = 4$ for various number of reflectors for the dataset in Figure \ref{fig:HouseholderComparisons}.}
	\label{fig:Plot_of_reflectors}
\end{figure}
In this section we provide experimental results to illustrate the representation capabilities of the proposed methods.

\subsection{Sparsely representing data}

The input data that we consider is taken from popular test images from the image processing literature (pirate, peppers, boat etc.). The test datasets $\mathbf{Y} \in \mathbb{R}^{64 \times N}$ consist of $8 \times 8$ non-overlapping patches with their means removed and normalized $\mathbf{Y} = \mathbf{Y}/255$. We choose to compare the proposed methods on image data since in this setting fast transforms that perform very well, like the Discrete Cosine Transform (DCT) for example, are available. Our goal is to provide Householder based dictionaries that perform well in terms of representation error with a small number of reflectors $m$ in their composition.

%
%
%
%
Table \ref{tb:Table1} shows the root mean squared error (RMSE) achieved by dictionaries trained on each test image separately and then used to sparsely represent those particular images. We show the performances of HQ$_m$--DLA and H$_m$--DLA for $m=6$ and $m=12$ reflectors. For perspective, we also show the performance achieved by the DCT on one hand and general (orthonormal and unconstrained) dictionary learning on the other -- we use Q--DLA and Stagewise K--SVD (SK--SVD) \cite{SKSVD}. For non-orthonormal dictionaries we use the OMP algorithm \cite{AKSVD} in the sparse reconstruction step. As expected, increasing the number of reflectors decreases RMSE in all cases. The best performing method of the ones proposed in this paper and shown in the table is H$_{12}$--DLA. The worse performance of this approach is achieved for the barb test image. To understand why we can see in Figure \ref{fig:LenaVsBarb} the eigenvalue distribution of the matrix $\mathbf{\tilde{Z}}$ from \eqref{eq:theFirstZ} for barb and peppers. As shown, most of the eigenvalues are close to (or exactly) zero. The difference comes when analyzing the negative eigenvalues which in the case of peppers are fewer and have larger magnitude than those of barb. We mention that for the barb test image the performance of Q--DLA is matched only by H$_{24}$--DLA. Table \ref{tb:Table1} shows on top the reference DCT and the proposed \textit{fast} dictionaries performance while the bottom shows the \textit{slower} dictionaries, including the H$_{32}$--DLA which generally performs slightly better even than Q--DLA. We would like to note here that the general dictionaries designed via K--SVD or SK--SVD do exhibit high mutual coherence in general, even though we do not construct overcomplete dictionaries. For example, the dictionary designed via SK--SVD and that reaches the best performance in terms of RMSE for the image peppers has mutual coherence over $0.9$, very high.

\def\arraystretch{1.2}
\begin{table*}[t]
	\begin{center}
		\caption{Speed-up provided by Householder based dictionaries as compared to the general orthonormal dictionaries and DCT  -- in this case a fast implementation, the Fast Cosine Transform (FCT) \cite{FCT}, is considered. We count the number of operations necessary to apply the dictionary as a direct and inverse operator, i.e., the computation of the correlations $\mathbf{D}^T\mathbf{y}$. We do not compare with the general sparse approximation methods like OMP since they are much slower -- they are at least $s$ times slower than an orthonormal dictionary, by \eqref{eq:NOMPCholesky}. The number of reflectors $m$ for which the complexity of the proposed dictionaries approximately coincides with Q--DLA and FCT is $m = 32$ and $m=3$ respectively.}\label{tb:Table0}
		\begin{tabular}{|c|c|c|c|c|c|c|c|c|c|c|c|c|c|c|}
			\hline
			Number of reflectors $m$ & 1 & 2 & 3 & 4 & 6 & 8 & 12 & 16 & 20 & 24\\ \hline
			speed-up $\rho_\text{Q--DLA}$ \eqref{eq:speedupFCT} & 32$\times$ &  16$\times$ &  11$\times$ &   8$\times$ &   5$\times$ &   4$\times$  &  3$\times$ &   2$\times$  &  1.6$\times$ & 1.3$\times$ \\ \hline
			speed-up $\rho_\text{FCT}$ \eqref{eq:speedupFCT} & 3$\times$ &   1.5$\times$ &   1$\times$ &   0.8$\times$ &   0.5$\times$  &  0.4$\times$  &  0.3$\times$  &  0.2$\times$  &  0.2$\times$  & 0.1$\times$ \\ \hline
		\end{tabular}
	\end{center}
\end{table*}

In the case of H$_m$--DLA we have tested two strategies to update the reflectors: sequential (in order of their index) and random. Since the difference between the two is negligible, the results shown use sequential update.

In Figure \ref{fig:IterationEvolution} we show the representation error evolution of the proposed algorithms and of Q--DLA with each iteration. The plot shows the effectiveness of the initialization procedures and the monotonically decrease in the objective function value. As expected, Q--DLA and SK--SVD perform best while QH$_m$--DLA the worse. Still, for the number of reflectors considered $m \in \{ 12, 32 \}$ the differences are not large. When we consider a larger number of reflectors like, $m=32$, we see that in all cases the RMSE is only slightly higher than that of Q--DLA. 

In Figure \ref{fig:HouseholderComparisons} we show the representation error for a dataset $\mathbf{Y}$ consisting of $N = 12288$ patches from several test images. For reference, we show again the DCT and Q--DLA representation performance. It is easy to see from the plot that the performance of the fixed transform is reached with a small number of reflectors $m$ (3 in both the cases of the proposed methods). When we increase the number of reflectors, H$_m$--DLA reaches the performance of Q--DLA for $m  =20$ while QH$_m$--DLA converges to a slightly worse result. As discussed in Section IV, we did expect QH$_m$--DLA to always perform worse than Q--DLA. Notice that for a small number of reflectors the performance of H$_m$--DLA and QH$_m$--DLA are very close suggesting that the extra orthogonal constraint is natural in this regime. The results are interesting when comparing with the references: it is clear that the dictionaries based on reflectors match the performance of Q--DLA for $m < n/2$ while they outperform the fixed DCT transform for $m \ll n$. This shows that a full orthonormal dictionary can be avoided without sacrificing performance.

In Figures \ref{fig:Plot_of_sparsity} and \ref{fig:Plot_of_reflectors} we show the eigenvalues of $\mathbf{\tilde{Z}}$ for Householder dictionaries created by H$_m$--DLA using the dataset described in Figure \ref{fig:HouseholderComparisons}. The eigenvalues are distributed similarly, independent of the choice of sparsity $s$ and number of reflectors $m$. In Figure \ref{fig:Plot_of_reflectors} notice that the choice of $m$ determines the number of negative eigenvalues with large magnitudes. As explained in Section IV this drives the reduction in the objective function of the Householder dictionary learning problem.

As seen, QH$_m$--DLA and H$_m$--DLA perform similarly. For best performance H$_m$--DLA is preferred but when the dictionary learning procedure is time critical QH$_m$--DLA is a better choice given the small loss in performance.

\begin{figure*}[t]
	\centering
	\includegraphics[width=0.66\textwidth]{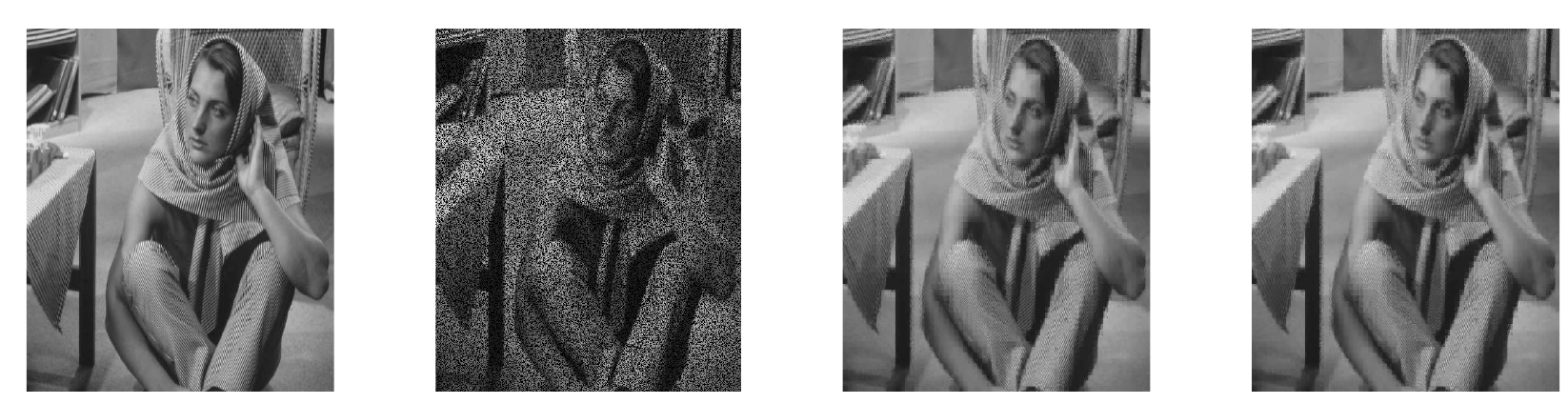}
	\caption{The figure contains from left to right: the original image, the corrupted image missing 40\% of the pixels chosen uniformly at random, the reconstruction using the orthonormal dictionary $(\text{MAE} = 0.0305, \text{MSE} = 0.0492)$ and the reconstruction using the Householder based dictionary with $m = 14$ reflectors $(\text{MAE} = 0.0321, \text{MSE} = 0.0512)$. We always have that $s=6$.}
	\label{fig:BarbNoisy}
\end{figure*}

\begin{figure*}[t]
	\centering
	\includegraphics[width=0.66\textwidth]{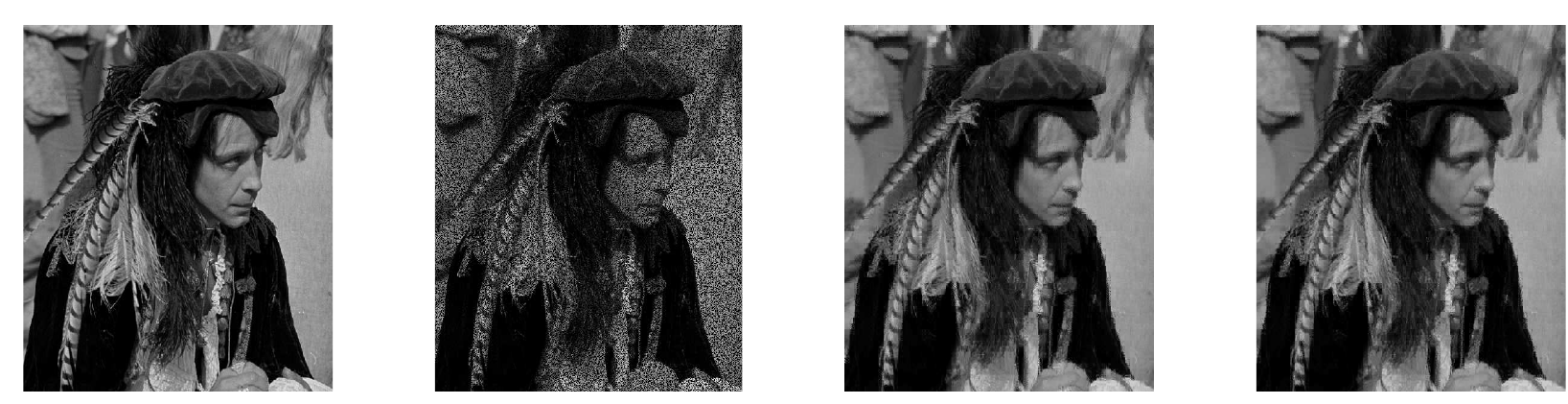}
	\caption{Analogous with Figure \ref{fig:BarbNoisy}. The orthonormal dictionary reaches $\text{MAE} = 0.0333, \text{MSE} = 0.0548$ and the Householder dictionary reaches $\text{MAE} = 0.0334, \text{MSE} = 0.0549$.}
	\label{fig:ManNoisy}
\end{figure*}

In Table \ref{tb:Table0} we show the speed-ups provided by the Householder based dictionaries as a function of the number of reflectors. We show the comparative computational complexity of using the dictionaries, not their training. We compare against the complexity of using a general orthonormal dictionary and against that of the DCT (we compare against an efficient implementation, the fast cosine transform). The cost of finding the largest entries in magnitude is the same for all methods and thus it is not accounted for. Since computing the correlations between the dictionary and a target signal takes $4nm$ for a Householder based dictionary with $m$ reflectors, the speed-ups are computed as
\begin{equation}
\rho_\text{Q--DLA} = \frac{(2n-1)n}{4nm},\ \rho_\text{FCT} = \frac{5/2 n \log n - 3n+6}{4nm}.
\label{eq:speedupFCT}
\end{equation}
The computational complexity of FCT is taken from \cite{FCT}. The latter is for perspective since it does not seem reasonable to assume that for image data we can construct a dictionary faster than the FCT that achieves the performance of Q--DLA. Still, notice that a Householder based dictionary with $m=3$ components closely matches the performance of the FCT both in terms of speed and in terms of performance (see Figure \ref{fig:HouseholderComparisons}). An important observation is that with $m=20$ reflectors we match closely the performance of Q--DLA while we still keep a computational advantage. From \eqref{eq:speedupFCT} it is clear that the proposed methods have lower computational complexity than general orthonormal dictionaries whenever $m \ll n$. We do not compare with the computational complexity of iterative methods since they are in general much slower than the methods discussed in this paper; for example, a batch variant of OMP called OMP--Cholesky \cite{AKSVD} needs
\begin{equation}
N_\text{OMP--Cholesky} = 2sn^2 + 2s^2 n + 4sn+s^3,
\label{eq:NOMPCholesky}
\end{equation}
operations. Since in general we do assume that we are dealing with sparse representations, i.e., $s \ll n$, the computational complexity of OMP--Cholesky is dominated by the first term which expresses the complexity of the explicit dictionary operator, the term that dominates also the computational complexity of using an orthonormal dictionary.

The final advantage of the proposed methods is the space requirement. As stated, in the case of dictionary learning the entries of the dictionaries need to be stored (or transmitted) together with the encoded data. With the proposed methods only the reflector vectors need to be stored, i.e., $mn$ entries. 

In terms of the computational complexity of the learning procedures themselves we report that in constructing the dictionaries for Figure \ref{fig:HouseholderComparisons} we have the approximate running times of 15 seconds for Q--DLA, 13 seconds for H$_8$--DLA, 7 seconds for QH$_8$--DLA all running for $K=100$ iterations while SK--SVD took over one minute. All running times include the initialization procedures. The simulations were conducted in the Mathworks Matlab$^\text{\textregistered}$ 2014 environment, using a modern laptop computer i7 processor, 16 GB RAM running Windows$^\text{\textregistered}$. As such, more efficient implementations are possible and the purpose of reporting the running times here is to provide a sense of the complexity of the learning procedure itself.

\subsection{Application: denoising images}

We also choose to test the trained dictionaries in reconstructions scenarios to fill in missing pixels from an image \cite{KSVD}. The experimental environment is as follows. We train a general orthonormal dictionary and one based on Householder reflectors on uncorrupted data (non-overlapping $8 \times 8$ image patches). We then blank a fixed percentage of the pixels in the images and perform the reconstruction using the previously trained dictionaries. Performance is measured in mean absolute error (MAE) and mean squared error (MSE) and the results are shown in Figures \ref{fig:BarbNoisy} and \ref{fig:ManNoisy}.

We compare Q--DLA and H$_{14}$--DLA to show that there are no large performance drawbacks when using dictionaries that are computationally efficient.

\section{Conclusions}
In this manuscript we describe algorithms for the orthonormal dictionary learning task based on Householder reflectors. We are able to construct dictionaries that can be efficiently manipulated and that also perform very well in terms of representation capabilities where we compare with the fast, fixed transforms and general orthonormal, learned dictionaries. We are also able to provide local minimum conditions for the Householder based and general orthonormal dictionary learning problems.

\section*{Acknowledgment}
The authors would like to thank anonymous reviewers and Bogdan Dumitrescu whose comments greatly improved the clarity of this manuscript.

\bibliographystyle{IEEEtran}
\bibliography{refs}

\end{document}